\documentclass{article}

\usepackage[preprint]{neurips_2023}

\usepackage[utf8]{inputenc} 
\usepackage[T1]{fontenc}    
\usepackage{hyperref}       
\usepackage{url}            
\usepackage{booktabs}       
\usepackage{amsfonts}       
\usepackage{nicefrac}       
\usepackage{microtype}      
\usepackage{xcolor}         

\usepackage{amsmath,amssymb,amsfonts}
\usepackage{algorithmic}
\usepackage{textcomp}
\usepackage{xcolor}
\usepackage{url}
\usepackage{graphicx}
\usepackage[subrefformat=parens]{subcaption}
\usepackage{multirow}
\usepackage{comment}
\usepackage{color}

\title{Action Q-Transformer: Visual Explanation in Deep Reinforcement Learning with Encoder-Decoder Model using Action Query}

\author{
    Hidenori Itaya\\
    Chubu University\\
    \texttt{itaya@mprg.cs.chubu.ac.jp}\\
    \And
    Tsubasa Hirakawa\\
    Chubu University\\
    \texttt{hirakawa@mprg.cs.chubu.ac.jp}\\
    \And
    Takayoshi Yamashita\\
    Chubu University\\
    \texttt{takayoshi@isc.chubu.ac.jp}\\
    \And
    Hironobu Fujiyoshi\\
    Chubu University\\
    \texttt{fujiyoshi@isc.chubu.ac.jp}\\
    \And
    Komei Sugiura\\
    Keio University\\
    \texttt{komei.sugiura@keio.jp} 
}

\begin{document}

\maketitle

\begin{abstract}
The excellent performance of Transformer in supervised learning has led to growing interest in its potential application to deep reinforcement learning (DRL) to achieve high performance on a wide variety of problems.
However, the decision making of a DRL agent is a black box, which greatly hinders the application of the agent to real-world problems. 
To address this problem, we propose the Action Q-Transformer (AQT), which introduces a transformer encoder-decoder structure to Q-learning based DRL methods.
In AQT, the encoder calculates the state value function and the decoder calculates the advantage function to promote the acquisition of different attentions indicating the agent's decision-making. 
The decoder in AQT utilizes action queries, which represent the information of each action, as queries.
This enables us to obtain the attentions for the state value and for each action.
By acquiring and visualizing these attentions that detail the agent's decision-making, we achieve a DRL model with high interpretability. 
In this paper, we show that visualization of attention in Atari 2600 games enables detailed analysis of agents' decision-making in various game tasks. 
Further, experimental results demonstrate that our method can achieve higher performance than the baseline in some games.
\end{abstract}

\section{Introduction}
In 2015, the Deep Q-Network (DQN), which introduces function approximation using a deep neural network (DNN) to Q-learning \cite{ql}, was proposed and achieved higher scores than human players in video game tasks \cite{dqn-nature}.
Subsequently, deep reinforcement learning (DRL), which combines deep learning and RL, has attracted much interest because it can handle a large number of states, such as images.
Also, since Transformer first appeared in the field of natural language processing in 2017 \cite{trans}, the Transformer structure has shown an impressive performance on a variety of tasks in the supervised learning field \cite{trans:vit,trans:detr}. 
Transformer enables longer-term memory of time-series data than the conventional convolutional neural network (CNN) and recurrent neural network (RNN).
As such, it is expected to be effective in the DRL field as well, and many methods that apply the Transformer structure to DRL have been proposed \cite{dt,mgdt}. 
DRL has achieved a high performance in a variety of tasks, including video games \cite{rl-game:shao2019}, robot control \cite{rl-robot:levine2018}, and self-driving \cite{rl-car:kiran2020}, because it can learn an agent's optimal behavior from a large state space in the environment.

DRL has achieved high performance on a variety of tasks, but there is the problem that the agent's decision-making is a black-box.
This is an obstacle when applying DRL agents in real-world applications because it relates to the reliability of the agents. 
There is a research field called explainable reinforcement learning (XRL) that aims to achieve models with high explainability and interpretability in DRL. 
Most of the prior XRL research has been done to interpret the model in the trained model, such as calculating the saliency map in the agent model. 
These methods require structural changes to the model at training, which may not be suitable for the explanations. 
Therefore, there is interest in methods that make the model structure highly interpretable in advance \cite{xrl:mott,xrl:pwn}.

In light of the above background, we propose Action Q-Transformer (AQT), a Q-learning-based DRL method with high interpretability for agents' decision-making. 
AQT utilizes a Transformer-encoder-decoder structure in which the encoder calculates state values and the decoder calculates advantages. 
The decoder acquires the attention for each action by using action queries that represent action information in the query. 
This allows us to acquire several different attentions to the agent's action choices and to achieve a detailed analysis of the agent's decision-making.

{\bf Contributions.}\quad
The main contributions of this paper are as follows.
\begin{itemize}
 \item 
    We propose a new Q-learning-based DRL method called Action Q-Transformer (AQT). 
    AQT acquires several attentions when calculating actions by introducing a Transformer encoder-decoder structure that utilizes action information as a query. In this way, we achieve a DRL model that is highly interpretable to the agent's decision-making.
\item 
    We introduce a Transformer structure in the agent model to achieve the same or improved performance compared to a baseline performance in the Atari 26000 benchmark.
\end{itemize}

\section{Related works}

\subsection{Deep reinforcement learning}
{\bf Q-learning-based DRL methods.}\quad
The Deep Q-Network (DQN) \cite{dqn-nature} delivers a higher performance than previous methods on video game tasks. 
DQN uses a neural network to represent the action value function Q $Q(a|s; \theta)$ and learns the network parameter $\theta$ to obtain the optimal action. 
Since its introduction, many DRL methods based on Q-learning \cite{ql} have been developed.

Hasselt {\it et al.} proposed Double DQN (DDQN), which uses two Q functions to calculate Q values, to solve the problem of overestimation in DQN \cite{ddqn}. 
DQN learns by randomly sampling acquired experience, which means it also samples unimportant experiences. 
Prioritized Experience Replay (PER) achieves efficient learning by prioritizing important experience \cite{per}. 
Because DQN learns by randomly sampling acquired experience, it also samples unimportant experience.
Prioritized Experience Replay (PER) achieves efficient learning by prioritizing important experience \cite{per}.
Wang {\it et al.} defined advantage, a value for the action itself, and proposed a Dueling network that decomposes the action value into state value and advantage \cite{dueling-network}. 
By estimating the state value and advantage separately, we can achieve learning that does not select the action if the advantage is low even if the state value is high.
The Q function in DQN is expressed as an expected value of return. 
In contrast, Bellemare {\it et al.} proposed distributional reinforcement learning, in which the Q function is represented as a distribution \cite{distribution}. 
This method improves the representation of action values and resolves unstable learning. 
The $\epsilon$-greedy method {\it etc.} implemented in DQN cannot sufficiently explore the state space of the environment.
Fortunato {\it et al.} proposed Noisy Networks, which add noise to the network that calculates actions, to achieve efficient exploration of the environment \cite{noisy-net}. 
As described above, many methods based on DQN with various innovations have been published. 
Hessel {\it et al.} proposed Rainbow, which combines these methods and achieved a high score on the Atari 2600 \cite{rainbow}. 
Since then, several improved methods have been developed and are among the most important techniques in DRL \cite{agent57,meme}.

{\bf Transformer in DRL.}\quad
Transformers have shown an impressive performance in various tasks in the supervised learning field, and the application of the Transformer structure has attracted interest in DRL as well.

Chen {\it et al.} proposed an off-line RL method called Decision Transformer (DT) that is based on the GPT architecture and uses returns as input \cite{dt}. 
The returns (reward, state, action) are of different data types, so the Transformer structure is applied to DRL by converting them into vectors using an embedding layer.
Many methods that introduce a Transformer structure focus on offline RL, but when applied to real problems, offline RL requires fine-tuning through task-specific interactions with the environment after pre-training the policy with a dataset, which takes a lot of time to learn.
Zheng {\it et al.} proposed Online DT (ODT), in which the deterministic policy in DT is replaced by a probabilistic policy \cite{odt}.
This method achieves high performance through fine-tuning by online RL.
Other studies have attempted to obtain a generalized policy from multiple large datasets. 
Lee {\it et al.} proposed Multi-Game DT (MGDT), which trains models on a variety of datasets consisting of both experts and non-experts \cite{mgdt}. 
This allows high performance to be achieved on multiple Atari 2600 games with a single set of parameters.

\subsection{Explainable reinforcement learning}
Explainable AI (XAI) is gaining interest for its potential to improve the reliability of AI systems. 
As such, it is fast becoming an important research area for solving the black-box problem in AI. 
DRL has similarly shown potential for solving this problem. 
Therefore, several studies have focused on analyzing the decision-making of agents.

Greydanus {\it et al.} calculate perturbation images by applying a Gaussian filter to the gradient of back-propagation in the action selection of an agent \cite{xrl:greydanus}. 
By acquiring a saliency map from these perturbed images, the agent's decision-making is analyzed.
Weitkamp {\it et al.} proposed a bottom-up method for the actor to compute the policy by applying a Grad-CAM \cite{xai:grad-cam} -based gazing area calculation method \cite{xrl:weitkamp2018}. 
However, these methods require back-propagation when analyzing agents' decision-making because they are bottom-up methods.
Manchin {\it et al.} introduced self-attention to PPO \cite{ppo}, aiming at improving the score as well as analyzing the policy \cite{xrl:manchin}. 
This method analyzes agents' decision-making by visualizing their attention to the policy.
Zhang {\it et al.} proposed attention-guided imitation learning (AGIL), which guides the gazing area of a network based on a person's gaze information \cite{xrl:agil}. 
They estimated the human gaze by supervised learning and utilized it to augment the input to guide the agent's gaze during action selection.
Rupprecht {\it et al.} focused on the importance of states that obtain significantly higher or lower rewards, and proposed a method for generating states of interest to trained DRL agents \cite{xrl:rupprecht}. 
This method trains the generative model to output states that the agent recognizes from the real environment, and the model is then utilized to analyze the decision-making of the agent.
Shi {\it et al.} proposed a self-supervised interpretable network (SSINet) that generates a fine-grained attention mask to highlight task-related information relevant to the agent's decision-making \cite{xrl:shi}. 
They generated an attention mask for the agent's decision-making by introducing SSINet in front of the agent's actor network.
Mott {\it et al.} obtained two types of attention (context, spatial) by using query-based attention \cite{xrl:mott}. 
This method obtains context/spatial-related attention by generating an attention query from the query network using the LSTM state at the previous time as input. 
In contrast, our method explicitly obtains attention for each action by generating an action query using a one-hot vector of actions and then using it to calculate the advantage.
Kenny {\it et al.} proposed the Prototype-Wrapper Network (PW-Net), which clarifies the inference process by introducing human-friendly prototypes into the network for action selection \cite{xrl:pwn}. 
They showed that PW-Net does not deteriorate performance compared to the black-box model. 
In contrast, our method clarifies the inference process by using a Transformer encoder-decoder structure and introducing an action query. 
The introduction of the Transformer improves the scores in some game tasks compared to the baseline.

\section{Action Q-Transformer}
Deep reinforcement learning agents have achieved high performance in various tasks. 
However, the basis for decisions on acquired actions is unclear, which prevents the application of deep reinforcement learning to the real world. 
In response, we propose the Action Q-Transformer (AQT), which introduces the Transformer encoder-decoder structure into the Q-learning-based deep reinforcement learning method.

\subsection{Overview of AQT structure}
\begin{figure}[tb]
    \centering
    \includegraphics[width=1.0\linewidth]{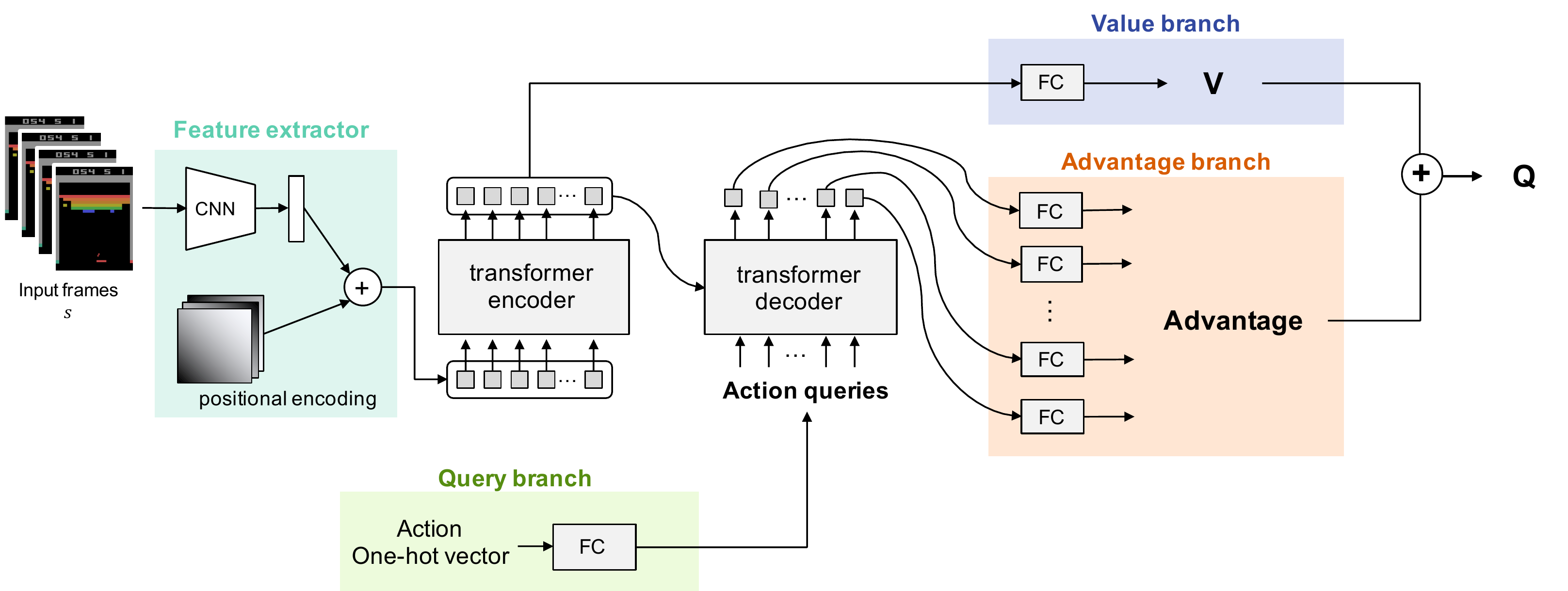}
    \caption{Overview of Action Q-Transformer structure.}
    \label{fig:aqt}
\end{figure}

AQT consists of a Feature extractor, Transformer Encoder-Decoder, Query branch, Value branch, and Advantage branch, as shown in Fig. \ref{fig:aqt}. 
AQT considers the relationship between patches in the input image at the encoder, and considers the relationship between the encoder output and action queries representing action information at the decoder.
Using each output branch, the encoder calculates the state value and the decoder calculates the advantage. 
Then, the action value for each action is calculated from the calculated state value and advantage. 
The agent's action in the environment is selected as the action with the largest calculated action value.

\begin{figure}[tb]
    \begin{tabular}{cc}
      \begin{minipage}[t]{0.44\hsize}
        \centering
        \includegraphics[width=1.0\linewidth]{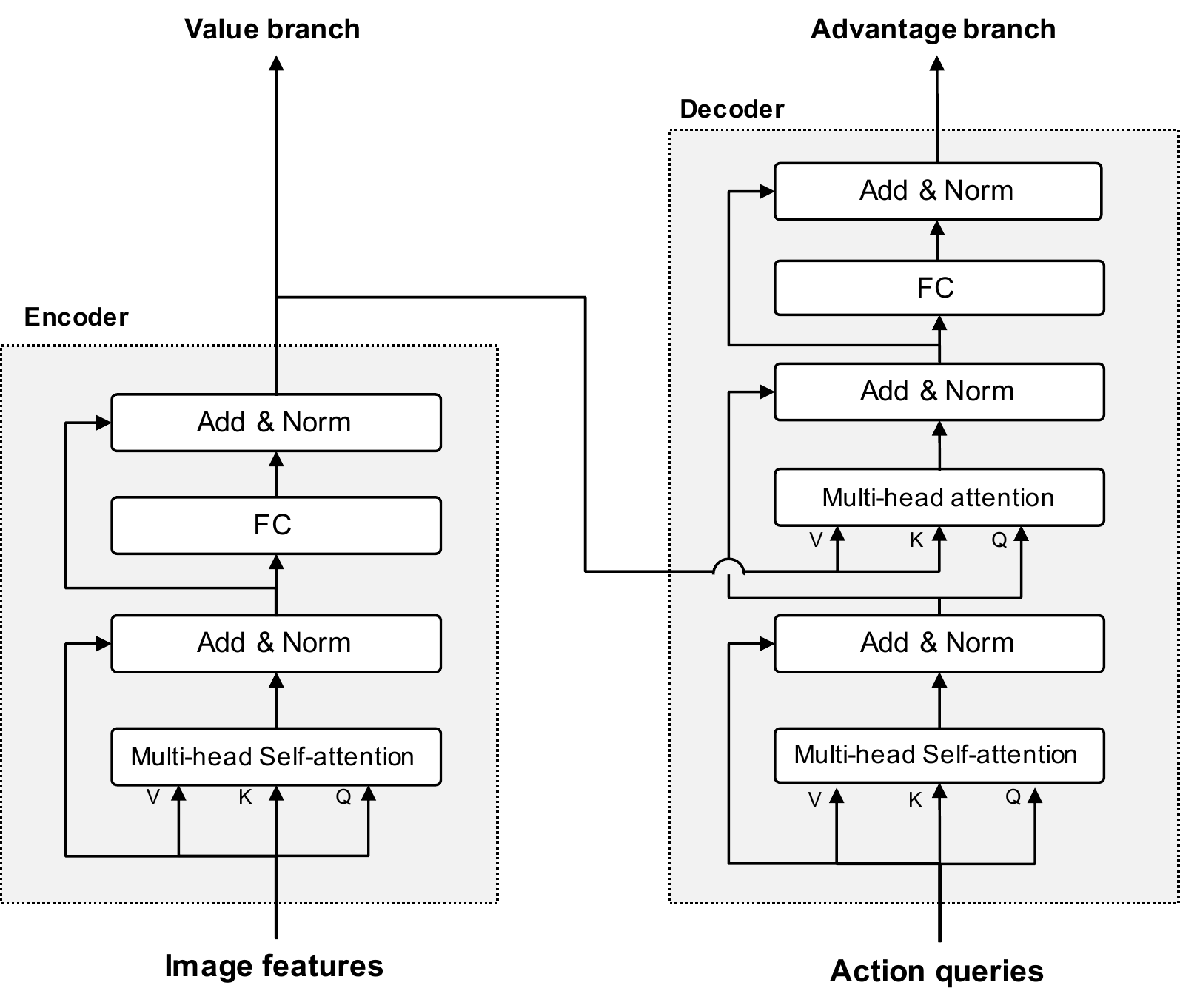}
        \subcaption{Transformer Encoder-Decoder.}
        \label{fig:aqt-ende}
      \end{minipage} &
      \hspace{0.02\columnwidth}
      \begin{minipage}[t]{0.44\hsize}
        \centering
        \includegraphics[width=1.1\linewidth]{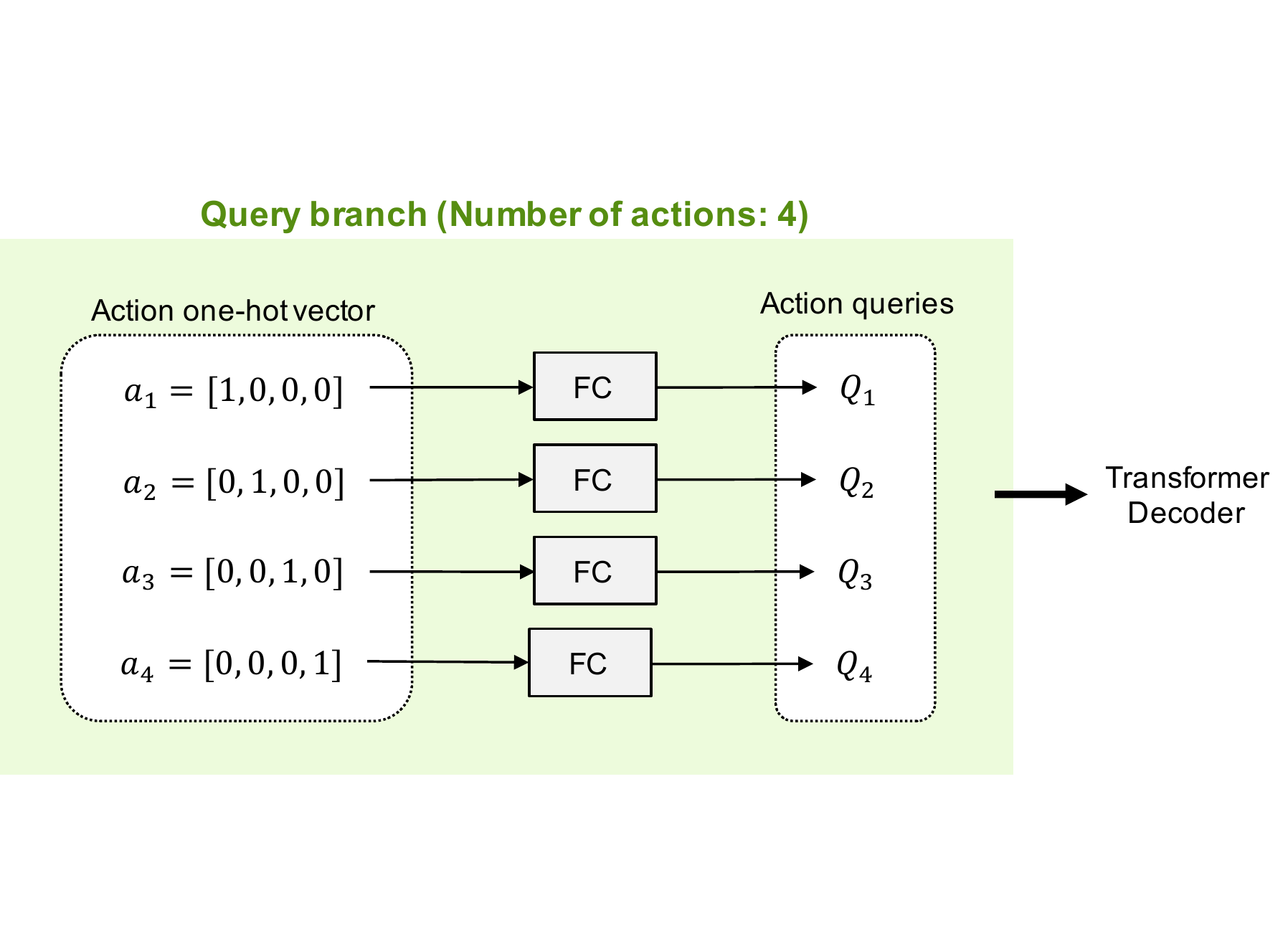}
        \subcaption{Query branch.}
        \label{fig:aqt-query}
      \end{minipage}
    \end{tabular}
    \caption{Details of Action Q-Transformer structure.}
\end{figure}

\subsection{Feature extractor}
The feature extractor obtains a feature map of a specific size from an input image using the CNN.
The acquired feature map is transformed into 1D features by adding location information to the feature map.
Each element in the 1D features corresponds to each element in the feature map.
In other words, each element of the feature map corresponds to a patch when the input image is divided into patches of the same size as the feature map.
The acquired 1D features are used as input values to the Transformer Encoder.

\subsection{Transformer Encoder-Decoder}
Our method implements a DRL model with high explainability for the agent's behavior by introducing a Transformer Encoder-Decoder, the structure of which is shown in Fig. \ref{fig:aqt-ende}.

{\bf Encoder-Decoder structure.}\quad
The encoder consists of a multi-head self-attention, a fully connected layer (FC), a residual mechanism, and a normalization process (Add \& Norm). 
Multi-head self-attention calculates the feature values by considering the relationship between elements in the feature map. 
Also, the output of the encoder is used as the input to the Value branch to obtain the attention of the state value of the input image from the encoder.

The decoder consists of four modules: the same module as the encoder, adding the Multi-head attention.
The value and key of the multi-head attention are the features acquired by encoder, and the query is the action queries acquired by the query branch (described below). 
The attention in the Decoder's Multi-head attention is calculated as
\begin{eqnarray}
\label{eq:decoder-att}
att_{de}(Q_{act},F_{en})={\rm softmax} \left( \frac{Q_{act} \cdot F_{en}^{T}}{\sqrt{dim}} \right)F_{en} ,
\end{eqnarray}
where $Q_{act}$ is the action queries after multi-headed self-attention, $F_{en}$ is the encoder output value, $dim$ is the dimensionality of $Q_{act}$ and $F_{en}$, and ${\rm softmax}(\cdot)$ is the softmax function.
The output of the decoder is used as input to the advantage branch to obtain attention for each action from the decoder.

By introducing this Transformer-encoder structure, the attention to state values and actions, which are important elements in understanding the agent's decision-making, is acquired. 
For action values, by using the decoder's Query as action information, it is possible to obtain the attention of each action and achieve a DRL model with high interpretability for the agent's decision-making.

{\bf Query branch.}\quad
In our method, advantage calculation for each action is achieved by using the action information as a query of the Transformer decoder. 
Then, the Query branch calculates the Action queries to be used for the decoder. 
Figure \ref{fig:aqt-query} shows an example of the Query branch when there are four agent actions. 
The process flow in the Query branch for this case is as follows. 
First, we define one-hot vectors for each action, assuming that the agent's actions are discrete. 
Next, the features of each one-hot vector are calculated by the FC layer and used as queries (action queries) for the decoder. 
The number of queries is the same as the number of actions, and the parameters of the FC layer used for feature extraction are updated during training, as in the case of other layers. 
This enables advantage calculations for each action using action queries, which are learnable parameters.

\subsection{Value and Advantage branches}
In the Value and Advantage branches, state value and advantage are calculated for each branch to obtain the final output, namely, the action value. 
The value branch uses the output value of the encoder and the advantage branch uses the output value of the decoder as input values for each branch. 
The action value $Q(s_{t},a)$ is calculated using state value $V(s_{t})$ and advantage $Adv(s_{t},a)$, as
\begin{eqnarray}
\label{eq:q-value}
Q(s_{t},a)=V(s_{t})+Adv(s_{t},a)-\frac{\sum_{i=1}^{n_{a}}Adv(s_{t},a_{i})}{n_{a}},
\end{eqnarray}
where $n_{a}$ is the number of agent actions.

\subsection{Training algorithm of Q-values in AQT}
\label{sec:aqt-algo}
The training algorithm in AQT uses Rainbow \cite{rainbow}.
However, since AQT introduces the transformer encoder-decoder structure, it is considered to have the same properties as the transformer in learning.
The performance of the transformer is known to have a scaling law that improves accuracy depending on the amount of data used for training \cite{scalinglaw}.
Therefore, we consider that the score improvement of AQT depends on the number of training steps.
To address this problem, we attempt to solve the above problem by introducing a loss function, Target Trained Q-network, in which the output of the learned baseline model is used as the target value.

{\bf Target Trained Q-network (TTQ).}\quad
The TTQ is designed as the squared error between the output $Q_{\rm aqt}$ of the AQT model and the target value, with the action value $Q_{\rm base}$ output by the learned baseline model as the target value. 
The loss function $L$ of AQT with TTQ $L_{\rm ttq}$ is calculated as
\begin{eqnarray}
\label{eq:ttq}
L &=& L_{\rm aqt} + \alpha L_{\rm ttq}, \nonumber \\
L_{\rm ttq} &=& (Q_{\rm base}(s_{t},a; \theta_{\rm base}) - Q_{\rm aqt}(s_{t},a; \theta_{\rm aqt}))^{2},
\end{eqnarray}
where $L_{\rm aqt}$ is the loss function similar to that of the algorithm used for training (Rainbow in our method), $\alpha$ is the learning rate of $L_{\rm ttq}$, and $\theta_{\rm base}, \theta_{\rm aqt}$ are the network parameters of each model.

\section{Experiments}
We conducted experiments using OpenAI gym game tasks to evaluate the effectiveness of AQT. 
In these experiment, we used 17 Atari 2600 games and analyzed the agents' decision-making by visualizing the attention acquired in ``Breakout'' and ``Seaquest''. (Visualization results for other games are shown in Appendix \ref{appendix:attention}.)
The compared methods are the baseline model ``Rainbow'' and our method ``AQT''. 
The end of training for each method is when the number of training steps reaches $5.0\times 10^{7}$. 
The episode ends either when one play of the game is completed or when the number of steps reaches $1.08\times 10^{5}$.  (Other hyper-parameters are shown in Appendix \ref{appendix:hypara}.)
The following two evaluation methods were used.
\begin{itemize}
    \item Score comparison on the Atari 2600
    \item Visual explanations using attention
\end{itemize}

\begin{figure}[tb]
    \centering
    \includegraphics[width=1.08\linewidth]{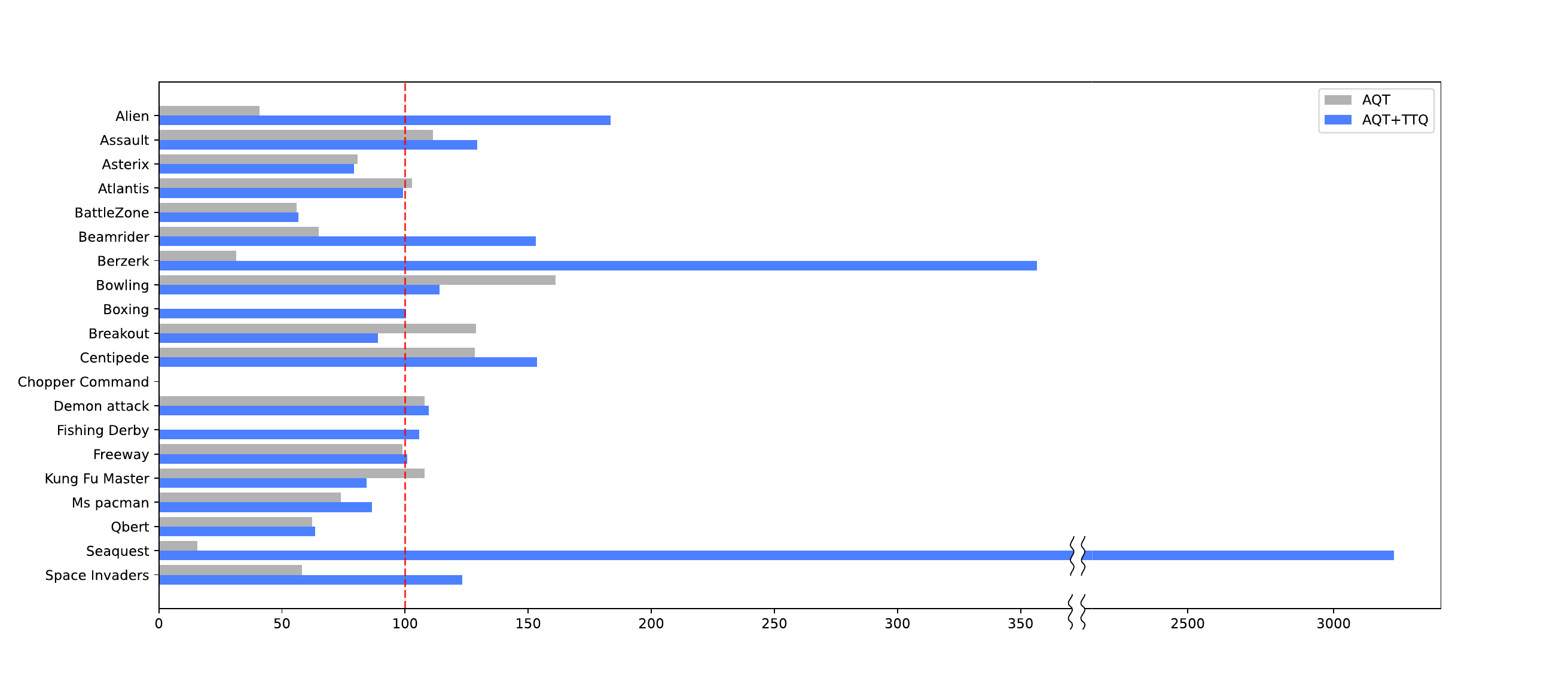}
    \caption{
    \textbf{Mean scores over 100 episodes on Atari 2600.}
    The graph shows the percentage of the average score of each method when the mean score of the baseline Rainbow model (red dashed line) is 100.
    }
    \label{fig:aqt-score}
\end{figure}

\subsection{Score comparison on the Atari 2600}
The mean scores for the 100 episodes on the Atari 2600 are shown in Fig. \ref{fig:aqt-score}. 
As we can see, AQT (gray) performs better than the baseline Rainbow (red dashed line) in games such as Bowling and Breakout. 
These are relatively simple games as they feature tasks with a low number of agent actions and a small environment search space. 
Therefore, we assume that AQT using the Transformer structure can accurately estimate the q-value of each action and thereby improve the score.
On the other hand, the AQT (gary) score improvement cannot be confirmed in Fishing derby and Seaquest {\it etc.}.
Games for which no improvement in AQT scores could be confirmed are more difficult to learn than games for which scores improved.
In such a task where learning is difficult, the Transformer scaling rule described in Sec.\ref{sec:aqt-algo} seems to have a particularly large impact.
Hence, we consider that the AQT scores are lower than the baseline Rainbow scores in game tasks that are difficult to learn, such as Seaquest.

The AQT with TTQ (AQT+TTQ) in our experiments is implemented by linearly decreasing the learning rate $\alpha$ of TTQ from $1.0$ to $0.0$ over $2.5\times 10^{7}$ steps in order to take into account the improvement in performance due to the loss function $L_{\rm aqt}$ of the AQT. (Experiments with various patterns of TTQ learning rate are discussed in Appendix \ref{appendix:ttq}.) 
In other words, the learning after $2.5\times 10^{7}$ steps is the same as that for AQT. 
The mean scores of AQT with TTQ (AQT+TTQ) over 100 episodes on the Atari 2600 are listed in Table \ref{fig:aqt-score}. 
As we can see, AQT+TTQ improves the scores compared to Rainbow in games where AQT was not able to do so (Beamrider, Fishing Derby, Seaquest, {\it etc.}). 
On the other hand, AQT without the TTQ scored higher in games (Bowling, Breakout, etc.) in which the score improvement was confirmed in AQT. 
This is because the target Rainbow score was lower than the AQT score, and the TTQ may have prevented learning on the AQT.

These results demonstrate that our method (AQT, AQT+TTQ) outperforms the baseline Rainbow. 
We also confirmed the effectiveness of the Transformer encoder-decoder structure.

\subsection{Visual explanations using Transformer attention}
\begin{figure}[t!]
\begin{tabular}{cc}
\begin{minipage}[t]{1.0\hsize}
\centering
    \includegraphics[clip,scale=0.46]{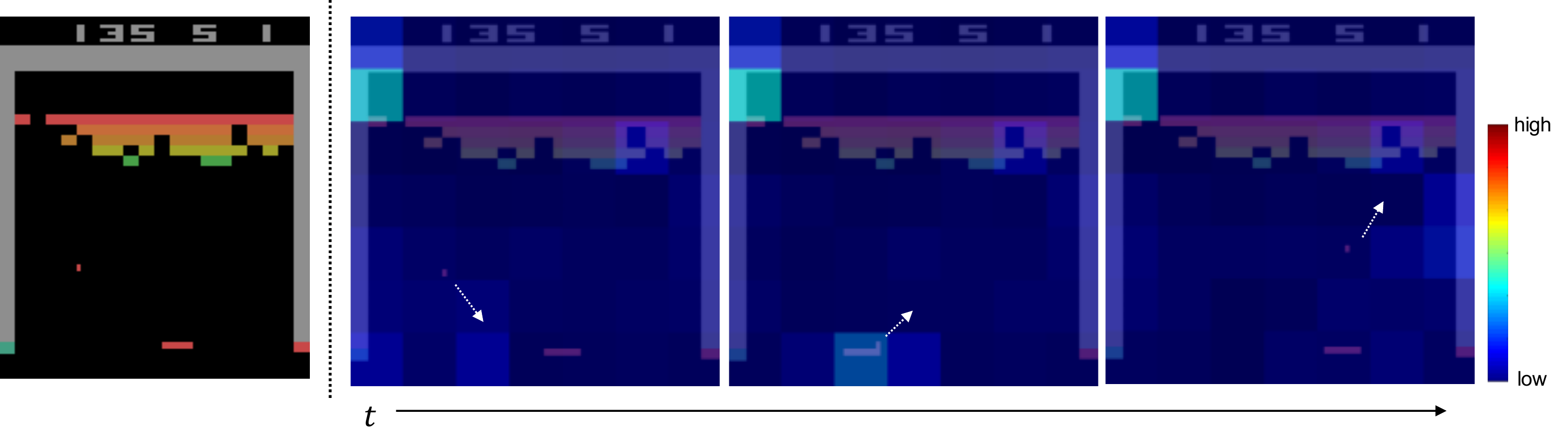}
\subcaption{\textbf{Encoder-attention.}: 
Example visualization of a scene in which a ball is hit back with a paddle. 
The visualization examples are arranged horizontally in time order. 
}
\vspace{0.2cm}
\label{fig:bo-encoder}
\end{minipage} \\
\begin{minipage}[t]{1.0\hsize}
\centering
    \includegraphics[width=0.98\linewidth]{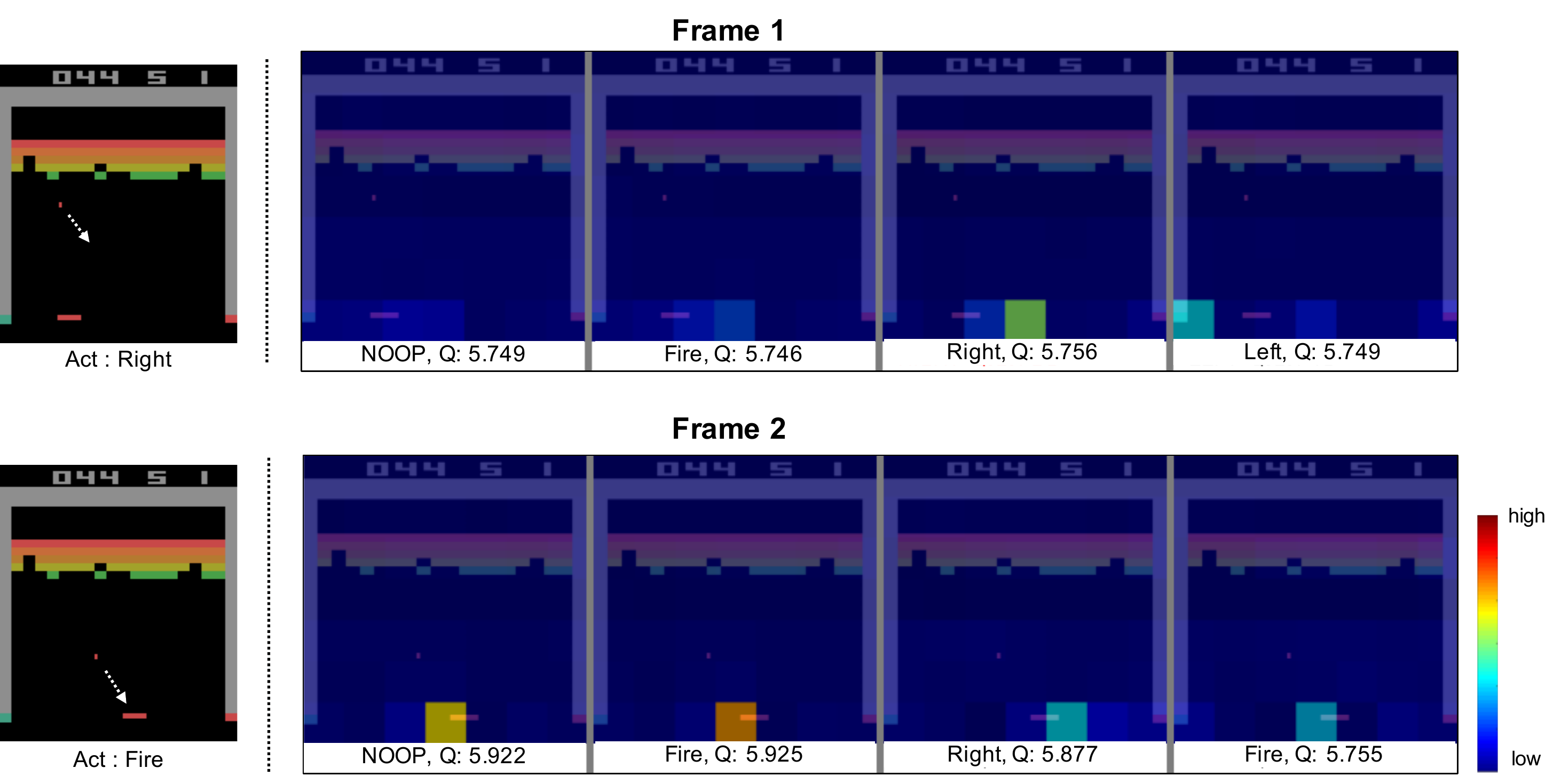}
\subcaption{\textbf{Decoder-attention.}: 
The caption below the attention indicates the name and Q-value of the action. 
The caption at the bottom of the raw image indicates the action selected by the agent in the current frame. 
``Fire'' in Breakout behaves the same as ``Noop''.
}
\label{fig:bo-decoder}
\end{minipage} \\
\end{tabular}
\caption{
\textbf{Visualization example of attention in Breakout.}
Arrows show the direction of travel of the ball.
}
\label{fig:visual-bo}
\end{figure}

\begin{figure}[t!]
\begin{tabular}{cc}
\begin{minipage}[t]{1.0\hsize}
\centering
    \includegraphics[clip,scale=0.44]{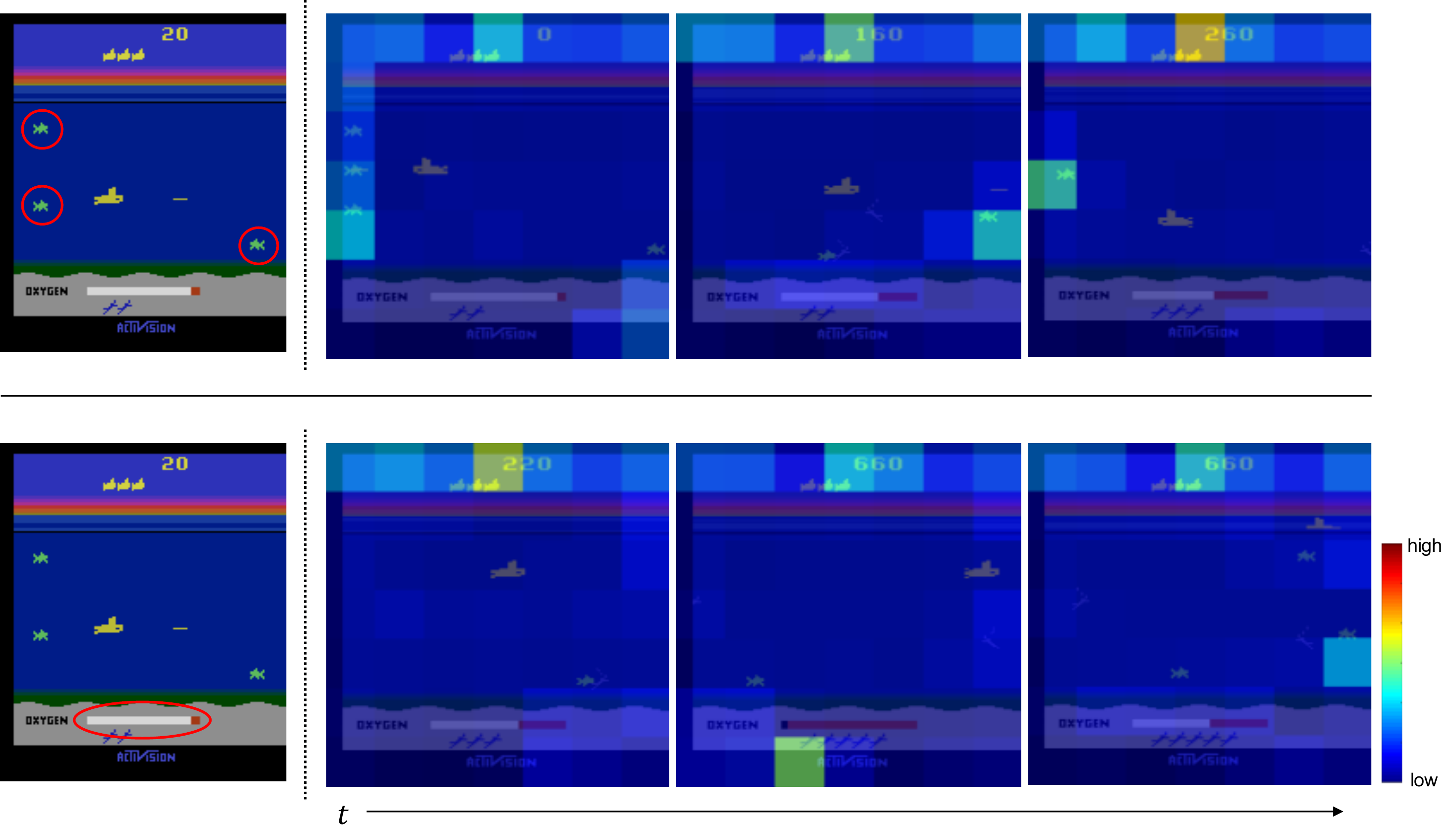}
\subcaption{\textbf{Encoder-attention.}: 
The top is an example of visualization for fish (red circles) in several frames.
The bottom is an example of visualization for an oxygen gauge (red circle) in a scene where oxygen has been depleted and replenished.
The visualization examples are arranged horizontally in time order.
}
\label{fig:sq-encoder}
\end{minipage} \\
\begin{minipage}[t]{1.0\hsize}
\centering
    \includegraphics[width=0.98\linewidth]{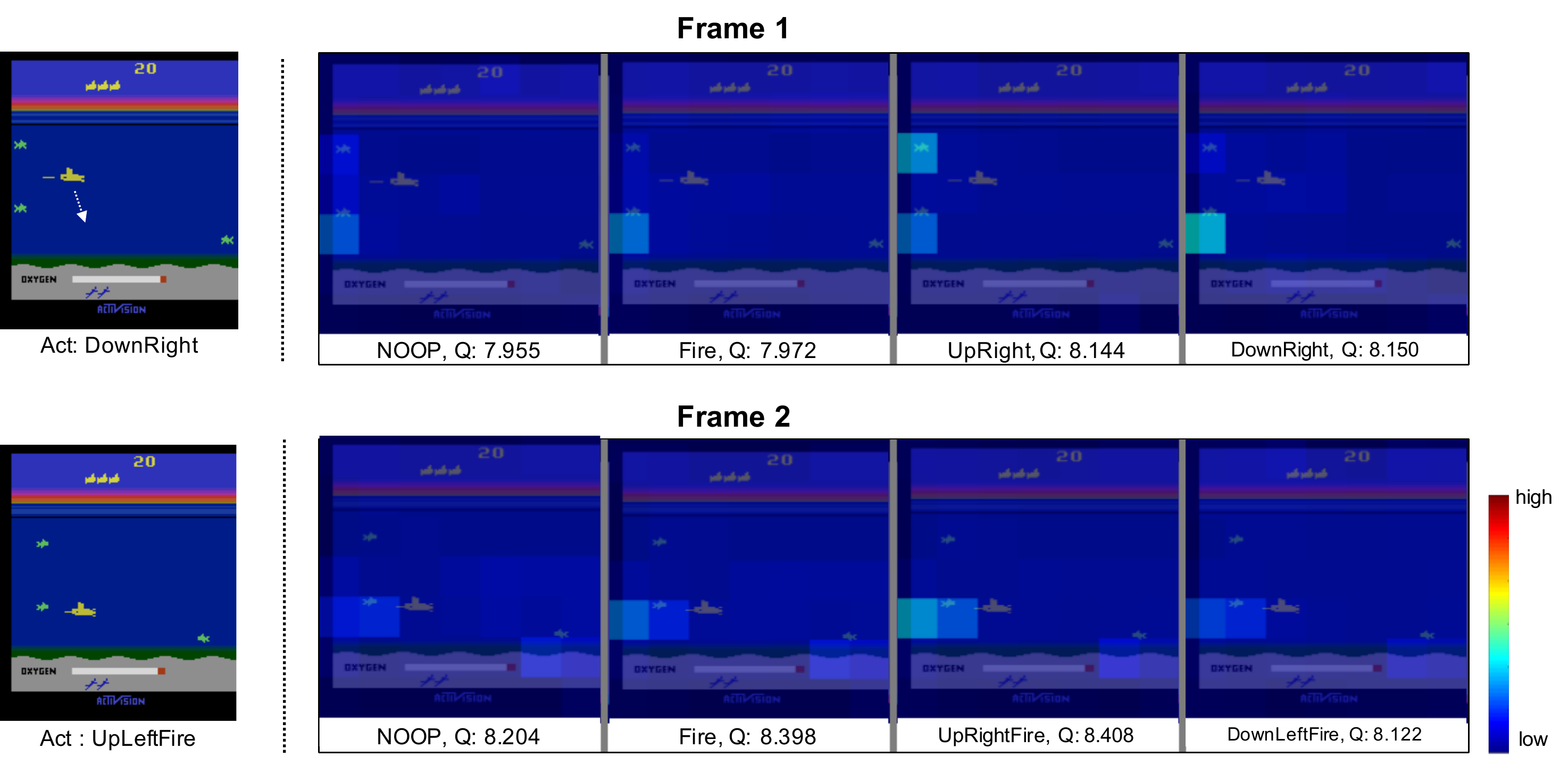}
\subcaption{\textbf{Decoder-attention.}: 
The caption below the attention indicates the name and Q-value of the action. 
The caption at the bottom of the raw image indicates the action selected by the agent in the current frame. 
The white arrow indicates the direction of the submarine. 
The total number of actions taken by Seaquest agents is 18, but 4 actions are picked up and shown here.
}
\label{fig:sq-decoder}
\end{minipage}
\end{tabular}
\caption{
\textbf{Visualization example of attention in Seaquest.}
}
\label{fig:visual-sq}
\end{figure}

Next, we visualized the attention using the AQT model for Breakout and the AQT+TTQ model for Seaquest, as shown in Figs. \ref{fig:visual-bo} and \ref{fig:visual-sq}, respectively. 
The model with the highest score in each game was used to visualize the attention (i.e., the AQT model for Breakout and the AQT+TTQ model for Seaquest). 
In the following, we explain the agent's decision making using the attention shown in these figures.

\subsubsection{Breakout}
Breakout is a game in which the player hits the ball back with the paddle to destroy the upper blocks.
There are four actions of the agent ($=$paddle): ``Noop'', ``Fire'', ``Left'', and ``Right''. 

{\bf Encoder attention (state value).}\quad
An example of encoder attention visualization of a scene in Breakout where a ball is hit back is shown in Fig. \ref{fig:bo-encoder}. 
Here, the agent is gazing at the paddle at the moment the ball is hit back and at the area with the fewest number of blocks. 
In Breakout, the score is obtained by destroying the block. 
Therefore, an important state for the agent is the state in which the ball is hit back with the paddle. 
In addition, when the agent makes a hole in a block and hits the ball back into it, the ball bounces on the top of the block and destroys many blocks. 
Based on the above, we believe that the agent correctly recognizes that the paddle and the ball are important objects at the moment of hitting the ball back, and that a hole in the block is important for obtaining a high score.

{\bf Decoder attention (advantage).}\quad
An example of visualization of decoder attention in Breakout is shown in Fig. \ref{fig:bo-decoder}. 
At Frame 1, the agent is gazing at the right side of the paddle for ``Right'' and at the left side of the paddle for ``Left''. 
In Breakout, the agent's action ``Right'' moves the paddle to the right and ``Left'' moves it to the left. 
Therefore, the agent is considered to be aware of the destination for each action. 
At Frame 2, the agent gazes at the paddle in all actions, and the higher the action value, the more strongly the agent gazes at the paddle. 
Frame 2 is a scene where the ball is approaching the paddle, so the agent needs to control the paddle to match the ball. 
Based on the above, we can assume that the more correctly the agent recognizes the destination of the paddle due to the action, and the more correctly the agent recognizes the destination of the ball in frames where the action leads to a score, the more correctly the agent recognizes the destination of the ball in frames where the action is to be selected.

\subsubsection{Seaquest}
Seaquest is a game in which the player must destroy enemies and fish while rescuing divers. The agent (=submarine) has 18 actions in total: 17 combinations of ``Fire'', ``Up'', ``Down'', ``Left'', and ``Right'', and an additional ``Noop''.

{\bf Encoder attention (state value).}\quad
An example of encoder attention visualization focusing on fish and oxygen gauges in Seaquest is shown in Fig. \ref{fig:sq-encoder}. 
At the top of the figure, the agent is gazing at the fish in all frames, and from the bottom, the agent is gazing at the oxygen gauge only for the frames in which it runs out of oxygen. 
In Seaquest, the score is obtained by destroying Fish. 
Also, the game is over when all the oxygen runs out, so the agents need to replenish their oxygen supply at regular intervals. 
Based on the above, we can assume that the agents recognize Fish as an important object in Seaquest and correctly recognize the need to replenish oxygen.

{\bf Decoder attention (advantage).}\quad
An example of visualization of decoder attention in Seaquest is shown in Fig.\ref{fig:sq-decoder}. 
From Frame 1, the agent is gazing at the upper-left Fish in ``UpRight'' and at the lower-left Fish in ``DownRight''. 
In Seaquest, the agent's action ``UpRight'' moves the submarine to the upper right and ``DownRight'' moves it to the lower right. 
Hence, the direction of movement of the submarine due to the agent's actions coincides with the position of Fish. 
Therefore, we can conclude that the agent is gazing at the Fish located at the destination. 
In Frame 2, which has moved to the lower right after ``DownRight'' was selected in Frame 1, ``UpLeftFire'' was selected, which includes an attack. 
In other words, in Frame 2, the agent is attacking the Fish it had gazed at in Frame 1. 
Based on the above, we consider that the agent recognizes Fish as a target to be destroyed and correctly identifies the most efficient target to attack for each action.

\subsubsection{Discussion}
By using the output of the encoder as input to the value branch, the encoder calculates attention on state values.
The state value represents the expected value of the return in the current state. 
Therefore, encoder attention indicates objects that are important for obtaining a score throughout all frames.
By using the output of the decoder as input to the advantage branch, the decoder computes attention on the agent's actions.
In our method, the decoder uses action queries to calculate attention for each action, rather than for all actions.
The advantage is the action value minus the state value, which is the value of action alone.
Therefore, the decoder attention indicates for each query (i.e., for each action) the objects or regions that would be affected if that action were selected.

\section{Conclusion}
In this paper, we proposed Action Q-Transformer (AQT), a Q-learning-based deep reinforcement learning method with high interpretability for agents' decision-making. 
AQT introduces an Transformer encoder-decoder structure in which the encoder calculates state values and the decoder calculates advantages. 
The structure of the decoder is such that it can acquire attention for each action by utilizing action queries that represent action information in the query. 
This enables the acquisition of different attentions regarding the agent's action choices and makes it possible to perform a detailed analysis of the agent's decision making.

Experiments using the Atari 2600 confirmed that the encoder can obtain attention to state value and that the decoder can obtain attention to actions. 
Moreover, we confirmed that the decoder can obtain attention to the query for each action, and that the region of attention differs for each action. 
By visualizing these different attentions that indicate agents' decision-making, we achieved an effective analysis for agents' decision-making in game tasks. 
Moreover, we showed that our method can obtain higher scores than a baseline thanks to providing a Target Trained Q-network (TTQ) as a solution to the Transformer scaling rule in the AQT. 
However, the experiments in this paper were conducted on a game task that is easy to analyze visually. 
Therefore, visual analysis of agents for more complex tasks (such as robot control and autonomous driving) using AQT remains a challenge for future work.

{
\small

\bibliography{arxiv_aqt}

\begin{thebibliography}{31}
\providecommand{\natexlab}[1]{#1}
\providecommand{\url}[1]{\texttt{#1}}
\expandafter\ifx\csname urlstyle\endcsname\relax
  \providecommand{\doi}[1]{doi: #1}\else
  \providecommand{\doi}{doi: \begingroup \urlstyle{rm}\Url}\fi

\bibitem[Badia et~al.(2020)Badia, Piot, Kapturowski, Sprechmann, Vitvitskyi,
  Guo, and Blundell]{agent57}
Badia, A.~P., Piot, B., Kapturowski, S., Sprechmann, P., Vitvitskyi, A., Guo,
  D., and Blundell, C.
\newblock {Agent57: Outperforming the atari human benchmark}.
\newblock In \emph{International conference on machine learning}, 2020.

\bibitem[Bellemare et~al.(2017)Bellemare, Dabney, and Munos]{distribution}
Bellemare, M.~G., Dabney, W., and Munos, R.
\newblock A distributional perspective on reinforcement learning.
\newblock In \emph{International Conference on Machine Learning}, 2017.

\bibitem[Carion et~al.(2020)Carion, Massa, Synnaeve, Usunier, Kirillov, and
  Zagoruyko]{trans:detr}
Carion, N., Massa, F., Synnaeve, G., Usunier, N., Kirillov, A., and Zagoruyko,
  S.
\newblock End-to-end object detection with transformers.
\newblock In \emph{European conference on computer vision}, pp.\  213--229.
  Springer, 2020.

\bibitem[Chen et~al.(2021)Chen, Lu, Rajeswaran, Lee, Grover, Laskin, Abbeel,
  Srinivas, and Mordatch]{dt}
Chen, L., Lu, K., Rajeswaran, A., Lee, K., Grover, A., Laskin, M., Abbeel, P.,
  Srinivas, A., and Mordatch, I.
\newblock Decision transformer: Reinforcement learning via sequence modeling.
\newblock \emph{Advances in neural information processing systems},
  34:\penalty0 15084--15097, 2021.

\bibitem[Dosovitskiy et~al.(2021)Dosovitskiy, Beyer, Kolesnikov, Weissenborn,
  Zhai, Unterthiner, Dehghani, Minderer, Heigold, Gelly, Uszkoreit, and
  Houlsby]{trans:vit}
Dosovitskiy, A., Beyer, L., Kolesnikov, A., Weissenborn, D., Zhai, X.,
  Unterthiner, T., Dehghani, M., Minderer, M., Heigold, G., Gelly, S.,
  Uszkoreit, J., and Houlsby, N.
\newblock An image is worth 16x16 words: Transformers for image recognition at
  scale.
\newblock In \emph{International Conference on Learning Representations}, 2021.

\bibitem[Fortunato et~al.(2018)Fortunato, Azar, Piot, Menick, Hessel, Osband,
  Graves, Mnih, Munos, Hassabis, Pietquin, Blundell, and Legg]{noisy-net}
Fortunato, M., Azar, M.~G., Piot, B., Menick, J., Hessel, M., Osband, I.,
  Graves, A., Mnih, V., Munos, R., Hassabis, D., Pietquin, O., Blundell, C.,
  and Legg, S.
\newblock Noisy networks for exploration.
\newblock In \emph{International conference on learning representations}, 2018.

\bibitem[Greydanus et~al.(2018)Greydanus, Koul, Dodge, and Fern]{xrl:greydanus}
Greydanus, S., Koul, A., Dodge, J., and Fern, A.
\newblock {Visualizing and understanding Atari agents}.
\newblock \emph{Proceedings of international conference on machine learning
  (ICML)}, 80:\penalty0 1792--1801, 2018.

\bibitem[Hessel et~al.(2018)Hessel, Modayil, Van~Hasselt, Schaul, Ostrovski,
  Dabney, Horgan, Piot, Azar, and Silver]{rainbow}
Hessel, M., Modayil, J., Van~Hasselt, H., Schaul, T., Ostrovski, G., Dabney,
  W., Horgan, D., Piot, B., Azar, M., and Silver, D.
\newblock {Rainbow: Combining improvements in deep reinforcement learning}.
\newblock In \emph{the AAAI conference on artificial intelligence}, 2018.

\bibitem[Kaplan et~al.(2020)Kaplan, McCandlish, Henighan, Brown, Chess, Child,
  Gray, Radford, Wu, and Amodei]{scalinglaw}
Kaplan, J., McCandlish, S., Henighan, T.~J., Brown, T.~B., Chess, B., Child,
  R., Gray, S., Radford, A., Wu, J., and Amodei, D.
\newblock Scaling laws for neural language models.
\newblock \emph{arXiv preprint, arXiv:2001.08361}, 2020.

\bibitem[Kapturowski et~al.(2023)Kapturowski, Campos, Jiang, Rakicevic, van
  Hasselt, Blundell, and Badia]{meme}
Kapturowski, S., Campos, V., Jiang, R., Rakicevic, N., van Hasselt, H.,
  Blundell, C., and Badia, A.~P.
\newblock Human-level atari 200x faster.
\newblock In \emph{International conference on learning representations}, 2023.

\bibitem[Kenny et~al.(2023)Kenny, Tucker, and Shah]{xrl:pwn}
Kenny, E.~M., Tucker, M., and Shah, J.
\newblock Towards interpretable deep reinforcement learning with human-friendly
  prototypes.
\newblock In \emph{International conference on learning representations}, 2023.

\bibitem[Kiran et~al.(2020)Kiran, Sobh, Talpaert, Mannion, Sallab, Yogamani,
  and P'erez]{rl-car:kiran2020}
Kiran, B.~R., Sobh, I., Talpaert, V., Mannion, P., Sallab, A. A.~A., Yogamani,
  S.~K., and P'erez, P.
\newblock Deep reinforcement learning for autonomous driving: A survey.
\newblock \emph{IEEE Transactions on Intelligent Transportation Systems},
  23:\penalty0 4909--4926, 2020.

\bibitem[Lee et~al.(2022)Lee, Nachum, Yang, Lee, Freeman, Xu, Guadarrama,
  Fischer, Jang, Michalewski, and Mordatch]{mgdt}
Lee, K.-H., Nachum, O., Yang, M., Lee, L.~Y., Freeman, D., Xu, W., Guadarrama,
  S., Fischer, I.~S., Jang, E., Michalewski, H., and Mordatch, I.
\newblock Multi-game decision transformers.
\newblock \emph{Advances in neural information processing systems},
  abs/2205.15241, 2022.

\bibitem[Levine et~al.(2018)Levine, Pastor, Krizhevsky, Ibarz, and
  Quillen]{rl-robot:levine2018}
Levine, S., Pastor, P., Krizhevsky, A., Ibarz, J., and Quillen, D.
\newblock {Learning Hand-Eye Coordination for Robotic Grasping with Deep
  Learning and Large-Scale Data Collection}.
\newblock \emph{The International Journal of Robotics Research}, 37\penalty0
  (4--5):\penalty0 421--436, 2018.

\bibitem[Manchin et~al.(2019)Manchin, Abbasnejad, and van~den
  Hengel]{xrl:manchin}
Manchin, A., Abbasnejad, E., and van~den Hengel, A.
\newblock {Reinforcement learning with attention that works: A self-supervised
  approach}.
\newblock In \emph{In International Conference on Neural Information
  Processing. Springer, Cham.}, pp.\  223--230, 2019.

\bibitem[Mnih et~al.(2015)Mnih, Kavukcuoglu, Silver, Rusu, Veness, Bellemare,
  Graves, Riedmiller, Fidjeland, Ostrovski, et~al.]{dqn-nature}
Mnih, V., Kavukcuoglu, K., Silver, D., Rusu, A.~A., Veness, J., Bellemare,
  M.~G., Graves, A., Riedmiller, M., Fidjeland, A.~K., Ostrovski, G., et~al.
\newblock {Human-level control through deep reinforcement learning}.
\newblock \emph{Nature}, 518\penalty0 (7540):\penalty0 529--533, 2015.

\bibitem[Mott et~al.(2019)Mott, Zoran, Chrzanowski, Wierstra, and
  Jimenez~Rezende]{xrl:mott}
Mott, A., Zoran, D., Chrzanowski, M., Wierstra, D., and Jimenez~Rezende, D.
\newblock {Towards interpretable reinforcement learning using attention
  augmented agents}.
\newblock \emph{Advances in neural information processing systems},
  32:\penalty0 12350--12359, 2019.

\bibitem[Rupprecht et~al.(2020)Rupprecht, Ibrahim, and J.Pal]{xrl:rupprecht}
Rupprecht, C., Ibrahim, C., and J.Pal, C.
\newblock {Finding and visualizing weaknesses of deep reinforcement learning
  agents}.
\newblock In \emph{International conference on learning representations}, 2020.

\bibitem[Schaul et~al.(2015)Schaul, Quan, Antonoglou, and Silver]{per}
Schaul, T., Quan, J., Antonoglou, I., and Silver, D.
\newblock Prioritized experience replay.
\newblock \emph{arXiv preprint arXiv:1511.05952}, 2015.

\bibitem[Schulman et~al.(2017)Schulman, Wolski, Dhariwal, Radford, and
  Klimov]{ppo}
Schulman, J., Wolski, F., Dhariwal, P., Radford, A., and Klimov, O.
\newblock {Proximal policy optimization algorithms}.
\newblock In \emph{arXiv preprint arXiv:1707.06347}, 2017.

\bibitem[Selvaraju et~al.(2017)Selvaraju, Cogswell, Das, Vedantam, Parikh, and
  Batra]{xai:grad-cam}
Selvaraju, R.~R., Cogswell, M., Das, A., Vedantam, R., Parikh, D., and Batra,
  D.
\newblock {Grad-CAM: Visual Explanations from Deep Networks via Gradient-based
  Localization}.
\newblock \emph{Proceedings of the IEEE international conference on computer
  vision}, pp.\  618--626, 2017.

\bibitem[Shao et~al.(2019)Shao, Tang, Zhu, Li, and Zhao]{rl-game:shao2019}
Shao, K., Tang, Z., Zhu, Y., Li, N., and Zhao, D.
\newblock {A survey of deep reinforcement learning in video games}.
\newblock \emph{arXiv preprint arXiv:1912.10944}, 2019.

\bibitem[Shi et~al.(2020)Shi, Huang, Song, Wang, Lin, and Wu]{xrl:shi}
Shi, W., Huang, G., Song, S., Wang, Z., Lin, T., and Wu, C.
\newblock Self-supervised discovering of interpretable features for
  reinforcement learning.
\newblock \emph{IEEE Transactions on Pattern Analysis and Machine
  Intelligence}, 2020.

\bibitem[Sutton \& Barto(1988)Sutton and Barto]{multistep}
Sutton, R.~S. and Barto, A.~G.
\newblock \emph{Reinforcement learning: An introduction}.
\newblock MIT press, 1988.

\bibitem[Van~Hasselt et~al.(2016)Van~Hasselt, Guez, and Silver]{ddqn}
Van~Hasselt, H., Guez, A., and Silver, D.
\newblock {Deep Reinforcement Learning with Double Q-Learning}.
\newblock In \emph{the AAAI conference on artificial intelligence}, 2016.

\bibitem[Vaswani et~al.(2017)Vaswani, Shazeer, Parmar, Uszkoreit, Jones, Gomez,
  Kaiser, and Polosukhin]{trans}
Vaswani, A., Shazeer, N., Parmar, N., Uszkoreit, J., Jones, L., Gomez, A.~N.,
  Kaiser, L.~u., and Polosukhin, I.
\newblock Attention is all you need.
\newblock In \emph{Proceedings of neural information processing systems},
  volume~30, 2017.

\bibitem[Wang et~al.(2016)Wang, Schaul, Hessel, Hasselt, Lanctot, and
  Freitas]{dueling-network}
Wang, Z., Schaul, T., Hessel, M., Hasselt, H., Lanctot, M., and Freitas, N.
\newblock {Dueling Network Architectures for Deep Reinforcement Learning}.
\newblock In \emph{International conference on machine learning}, pp.\
  1995--2003, 2016.

\bibitem[Watkins \& Dayan(1992)Watkins and Dayan]{ql}
Watkins, C.~J. and Dayan, P.
\newblock {Q-Learning}.
\newblock \emph{Machine learning}, 8\penalty0 (3-4):\penalty0 279--292, 1992.

\bibitem[Weitkamp et~al.(2018)Weitkamp, van~der Pol, and
  Akata]{xrl:weitkamp2018}
Weitkamp, L., van~der Pol, E., and Akata, Z.
\newblock {Visual rationalizations in deep reinforcement learning for atari
  games}.
\newblock In \emph{Benelux conference on artificial intelligence}, pp.\
  151--165, 2018.

\bibitem[Zhang et~al.(2018)Zhang, Liu, Zhang, Whritner, Muller, Hayhoe, and
  Ballard]{xrl:agil}
Zhang, R., Liu, Z., Zhang, L., Whritner, J.~A., Muller, K.~S., Hayhoe, M.~M.,
  and Ballard, D.~H.
\newblock {AGIL: Learning Attention from Human for Visuomotor Tasks}.
\newblock In \emph{Proceedings of the European conference on computer vision},
  pp.\  663--679, 2018.

\bibitem[Zheng et~al.(2022)Zheng, Zhang, and Grover]{odt}
Zheng, Q., Zhang, A., and Grover, A.
\newblock Online decision transformer.
\newblock In \emph{Proceedings of international conference on machine
  learning}, pp.\  27042--27059, 2022.

\end{thebibliography}
\bibliographystyle{icml2023}

\clearpage

\part*{Appendix}
This document is an appendix to ``Action Q-Transformer: Visual Explanation in Deep Reinforcement Learning with Encoder-Decoder Model using Action Query".

\setcounter{section}{0}
\renewcommand{\thesection}{\Alph{section}}

\section{Details of hyper-parameter settings and model structure in the our experiment}
\label{appendix:hypara}
The details of the hyper-parameter of each methods (The baseline Rainbow \cite{rainbow}, our method Action Q-Transformer (AQT)) used in the experiments in the our paper are shown in Table \ref{tab:hypara}.
The details of the structure of the Rainbow and AQT models are also shown in Table \ref{tab:rainbow_model},\ref{tab:aqt_model}.

\begin{table}[h!]
    \centering
    \caption{
        \textbf{Hyper-parameter settings in our experiments.}
        The upper shows the preprocessing of environmental information in Atari 2600.
        The lower shows the various parameters used at training.
        ``atoms" are the bins of the categorical distribution in Distributional RL.
        ``min/max values" are the minimum and maximum values of the categorical distribution.
    }
    \vspace{0.2cm}
    \scalebox{0.80}{
    \renewcommand{\arraystretch}{1.5}
    \begin{tabular}{c|c|c} \hline
        \multicolumn{2}{c|}{Hyper-parameter} & Value \\ \hline \hline
        \multirow{5}{*}{Environment} & Grey-scaling & True \\ \cline{2-3}
        & observation down-sampling & $(84, 84)$ \\ \cline{2-3}
        & frames stacked  & 4 \\ \cline{2-3}
        & action repetitions & 4 \\ \cline{2-3}
        & max steps per episode & $1.08 \times 10^{5}$ \\ \hline \hline
        \multirow{3}{*}{RL parameter} & optimizer, learning rate, $\epsilon$ & Adam, $6.25 \times 10^{-5}$, $1.5 \times 10^{-4}$ \\ \cline{2-3}
        & minibatch size & 32 \\ \cline{2-3}
        & discount factor & 0.99 \\ \hline
        \multirow{3}{*}{Prioritized experience replay \cite{per}} & memory size & $1.0 \times 10^{6}$ \\ \cline{2-3}
        & exponent & 0.5 \\ \cline{2-3}
        & initial importance sampling & 0.4 \\ \hline
        Reward clipping \cite{dqn-nature} & clipping range & [-1, 1] \\ \hline
        Target Q-network \cite{dqn-nature} & target network period & $8.0 \times 10^{3}$ \\ \hline
        Multi-step learning \cite{multistep} & multi-step returns & 3 \\ \hline
        Noisy network \cite{noisy-net} & initial standard deviation & 0.1 \\ \hline
        \multirow{2}{*}{Distributioanl RL \cite{distribution}} & atoms & 51 \\ \cline{2-3}
        & min/max values & [-10, 10] \\ \hline
    \end{tabular}
    }
    \label{tab:hypara}
\end{table}

\begin{table}[h!]
    \centering
    \caption{
        \textbf{Detail structures of Rainbow model.}
        The Rainbow model is constructed with 3 convolutional layers and each output branches that 2 liner layers implementing the noisy network.
        ``$N_{act}$" is the number of actions of the agent, and ``atoms" is the bins of the category distribution in Distributional RL (see Table \ref{tab:hypara} for the values).
    }
    \vspace{0.2cm}
    \scalebox{0.80}{
    \renewcommand{\arraystretch}{1.5}
    \begin{tabular}{c|c|c} \hline
        \multicolumn{2}{c|}{module} & contents \\ \hline \hline
        \multicolumn{2}{c|}{Conv. layer 1} & (channel, filter size, stride) = (32, $8 \times 8$, 4) \\ \hline
        \multicolumn{2}{c|}{Conv. layer 2} & (channel, filter size, stride) = (64, $4 \times 4$, 2) \\ \hline
        \multicolumn{2}{c|}{Conv. layer 3} & (channel, filter size, stride) = (64, $3 \times 3$, 1) \\ \hline
        \multirow{2}{*}{Value branch} & NoisyLinear 1 & (in, out) = (3136, 512) \\ \cline{2-3}
        & NoisyLinear 2 & (in, out) = (512, atoms) \\ \hline
        \multirow{2}{*}{Advantage branch} & NoisyLinear 1 & (in, out) = (3136, 512) \\ \cline{2-3}
        & NoisyLinear 2 & (in, out) = (512, atoms $\times N_{act}$)  \\ \hline
    \end{tabular}
    }
    \label{tab:rainbow_model}
\end{table}

\begin{table}[h!]
    \centering
    \caption{
        \textbf{Detail structures of AQT model.}
        The feature extractor is constructed with 3 convolutional layers, the query branch with 1 liner layer, and each output branches with 2 liner layers implementing the noisy network.
        ``$N_{act}$" is the number of actions of the agent, and ``atoms" is the bins of the category distribution in Distributional RL (see Table \ref{tab:hypara} for the values).
        ``hidden\_dim" is $N_{act} \times 32$.
    }
    \vspace{0.2cm}
    \scalebox{0.8}{
    \renewcommand{\arraystretch}{2}
    \begin{tabular}{c|c|c} \hline
        \multicolumn{2}{c|}{module} & contents \\ \hline \hline
        \multirow{3}{*}{Feature Extractor} & Conv. layer 1 & (channel, filter size, stride) = (32, $8 \times 8$, 4) \\ \cline{2-3}
        & Conv. layer 2 & (channel, filter size, stride) = (64, $4 \times 4$, 2) \\ \cline{2-3}
        & Conv. layer 3 & (channel, filter size, stride) = (hidden\_dim, $3 \times 3$, 1) \\ \hline
        \multicolumn{2}{c|}{Trasformer Encoder} & (in, head\_num, dim\_feedforward) = (hidden\_dim, 4, 64)\\ \hline
        Query branch & Liner & (in, out) = ($N_{act}$, hidden\_dim) \\ \hline
        \multicolumn{2}{c|}{Trasformer Decoder} & (in, head\_num, dim\_feedforward) = (hidden\_dim, 4, 64) \\ \hline
        \multirow{2}{*}{Value branch} & NoisyLinear 1 & (in, out) = ($7 \times 7 \times$ hidden\_dim, 512) \\ \cline{2-3}
        & NoisyLinear 2 & (in, out) = (512, atoms) \\ \hline
        \multirow{2}{*}{Advantage branch} & NoisyLinear 1 & (in, out) = (hidden\_dim, 512) \\ \cline{2-3}
        & NoisyLinear 2 & (in, out) = (512, atoms)  \\ \hline
    \end{tabular}
    }
    \label{tab:aqt_model}
\end{table}

\section{Experiment on learning rate for Target Trained Q-network}
\label{appendix:ttq}
In a score comparison on the Atari 2600, no score improvement was observed for AQT and AQT introducing Target Trained Q-network (TTQ) (AQT+TTQ) in some games (see Fig. 3 of the our paper).
Therefore, we conducted a similar experiment using 6 patterns of TTQ learning rate $\alpha$ designs for AQT+TTQ.
In this experiment, we used Atari 2600's ``Chopper Command" and ``Kung Fu Master".

The following 6 designs for TTQ learning rate are shown.
\begin{itemize}
\item 
AQT+TTQ: Decreases linearly with $2.5 \times 10^{7}$ steps from 1.0 to 0.0 (same as AQT+TTQ in the our paper)
\item 
AQT+TTQ $\alpha=0.2$: fixed at 0.2
\item 
AQT+TTQ $\alpha=0.4$: fixed at 0.4
\item 
AQT+TTQ $\alpha=0.6$: fixed at 0.6
\item 
AQT+TTQ $\alpha=0.8$: fixed at 0.8
\item 
AQT+TTQ $\alpha=1.0$: fixed at 1.0
\end{itemize}

The mean score between 100 episodes for each learning rate pattern is shown in Fig. \ref{fig:ttq-score}.
We can see the score improvement in Chopper Command by changing the design of the learning rate for TTQ.
Chopper Command is a task for which no score was obtained in AQT and AQT+TTQ.
In contrast, we don't see any score improvement in Kung Fu Master.
Kung Fu Master is the task that the AQT score was able to exceed the Rainbow score.
From these results, we can consider that the introduction of TTQ is effective for tasks in which AQT are not good at, and that the learning rate of TTQ needs to be optimally designed for each task.

\begin{figure}[h!]
    \centering
    \includegraphics[width=1.0\linewidth]{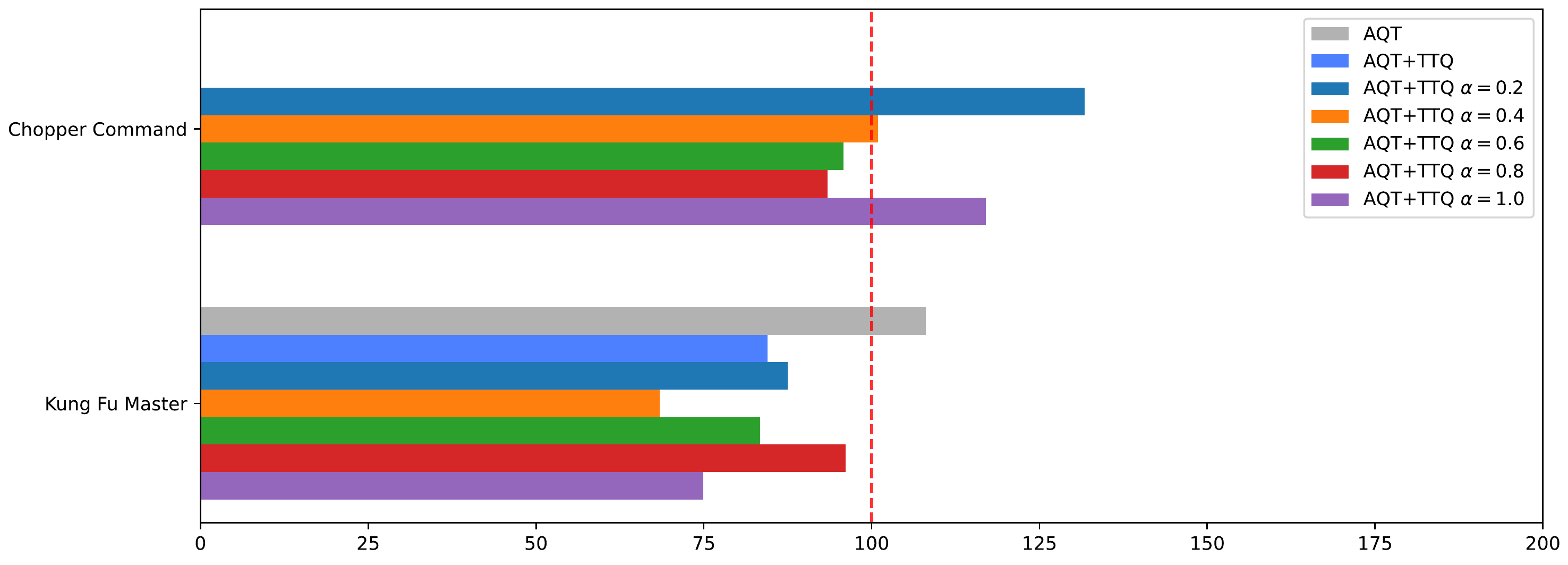}
    \caption{{\bf Mean score of 100 episodes on the Atari 2600 for the TTQ learning rate $alpha$.}
    The graph shows the percentage of the average score of each method when the mean score of the baseline Rainbow model (red dashed line) is 100.
    }
    \label{fig:ttq-score}
\end{figure}

\newpage

\section{Experimental results on Atari 2600}
\label{appendix:attention}
The learning curve of AQT for each Atari 2600 game is shown in Fig. \ref{fig:atari-graph}.
Visualization examples of Decoder attention other than Breakout and Seaquest are shown in Fig. \ref{fig:visual-decoder}.
Fig. \ref{fig:visual-decoder} shows examples of visualization of Kung Fu Master and Boxing.

\begin{figure}[h!]
    \centering
    \includegraphics[width=1.0\linewidth]{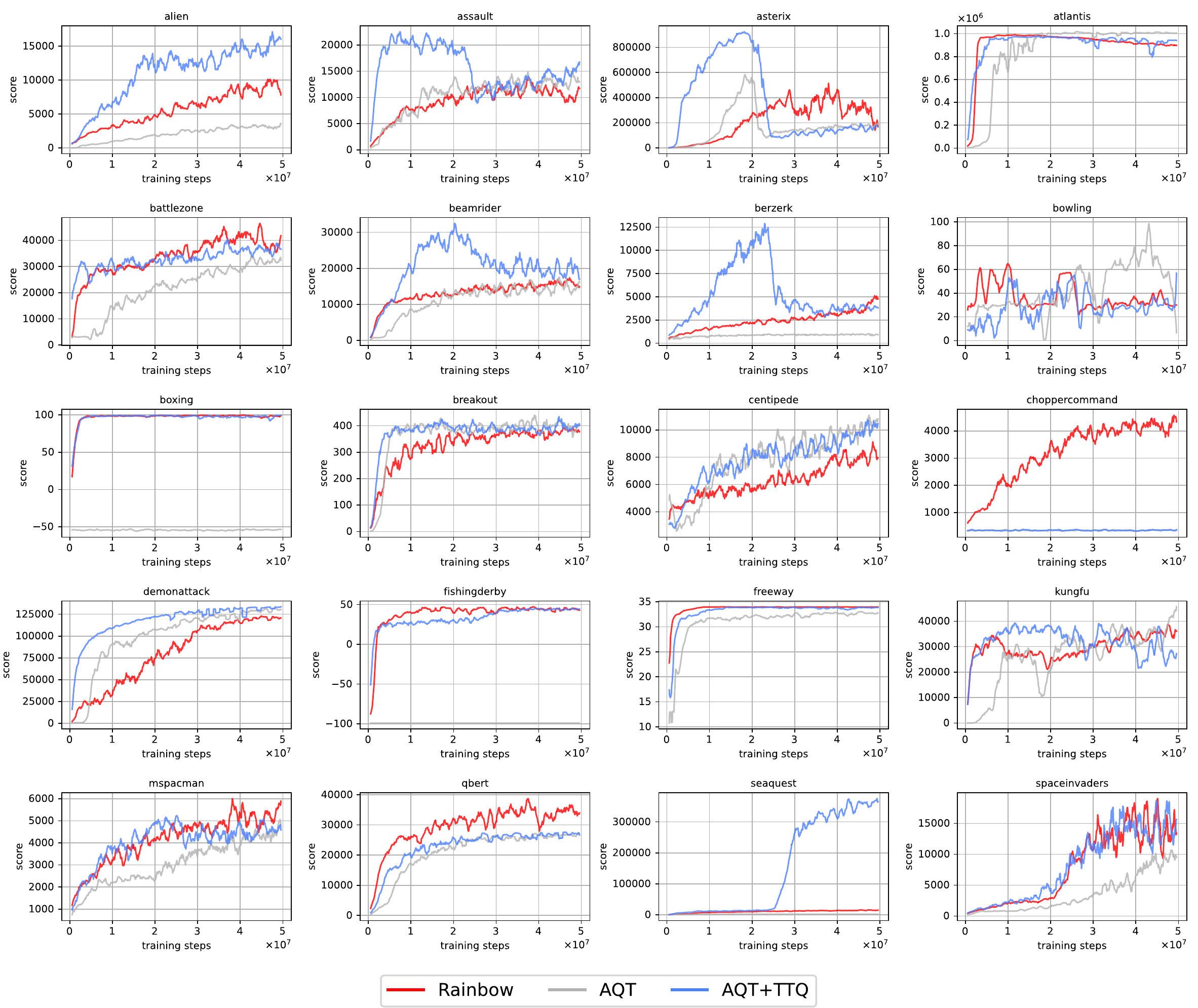}
    \caption{{\bf Learning curves for each method on Atari 2600.}
    The learning curves for the 20 Atari 2600 games described in the our paper are shown for each game.
    The graphs are smoothed using moving average.
    }
    \label{fig:atari-graph}
\end{figure}

\begin{figure}[t]
\begin{tabular}{cc}
\begin{minipage}[t]{1.0\hsize}
\centering
    \includegraphics[width=0.96\linewidth]{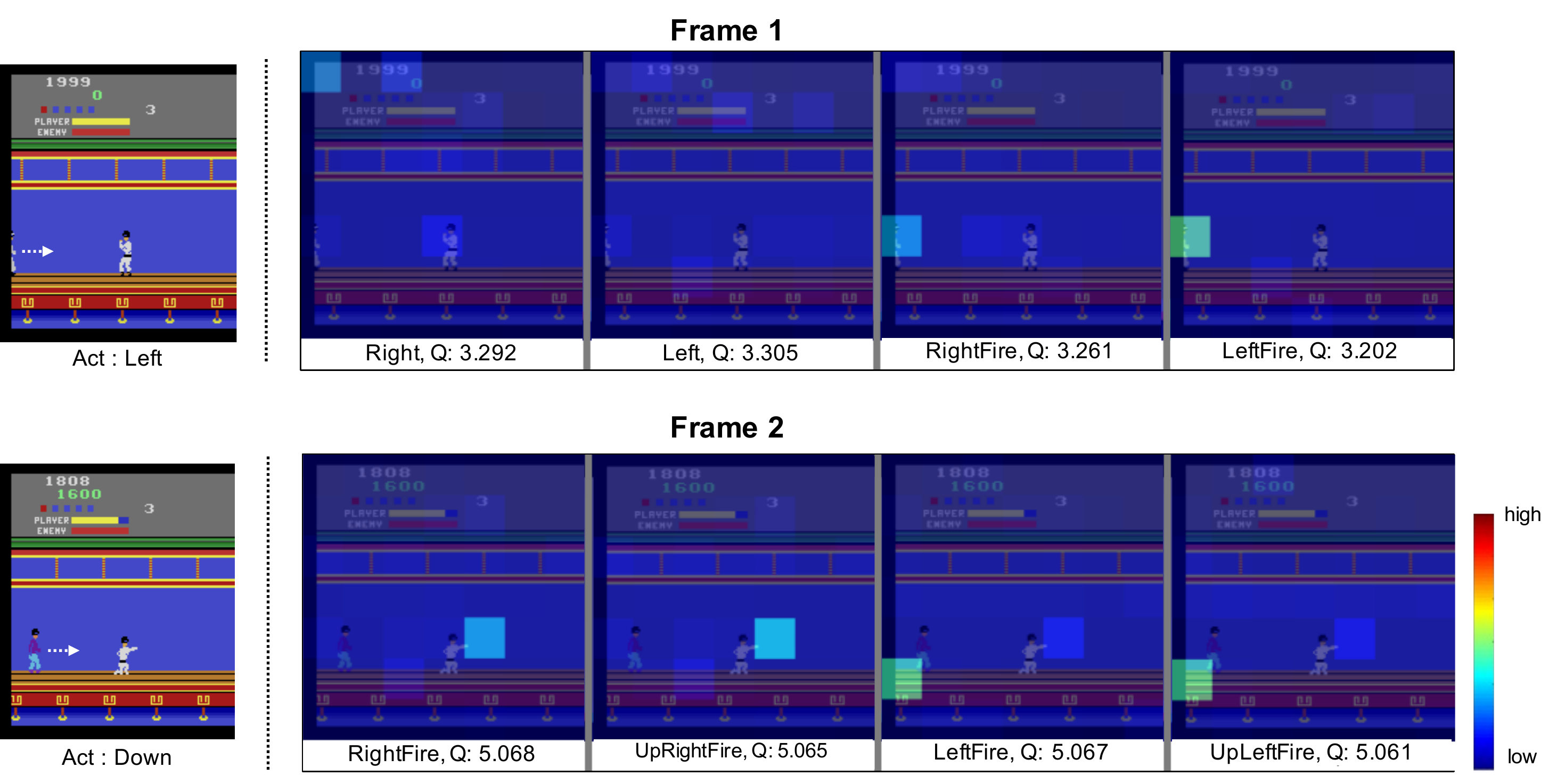}
\subcaption{\textbf{Kung Fu Master.}: 
White arrows indicate the direction in which the enemy is traveling.
The number of actions of Kung Fu Master’s agents is 18, but 4 actions are picked up and shown here.
Frame 1 is the scene where the enemy appears from the left side of the screen, and Frame 2 is the scene where the enemy is close to the player.
}
\vspace{0.2cm}
\label{fig:visual-kungfu}
\end{minipage} \\
\begin{minipage}[b]{1.0\hsize}
\centering
    \includegraphics[width=0.96\linewidth]{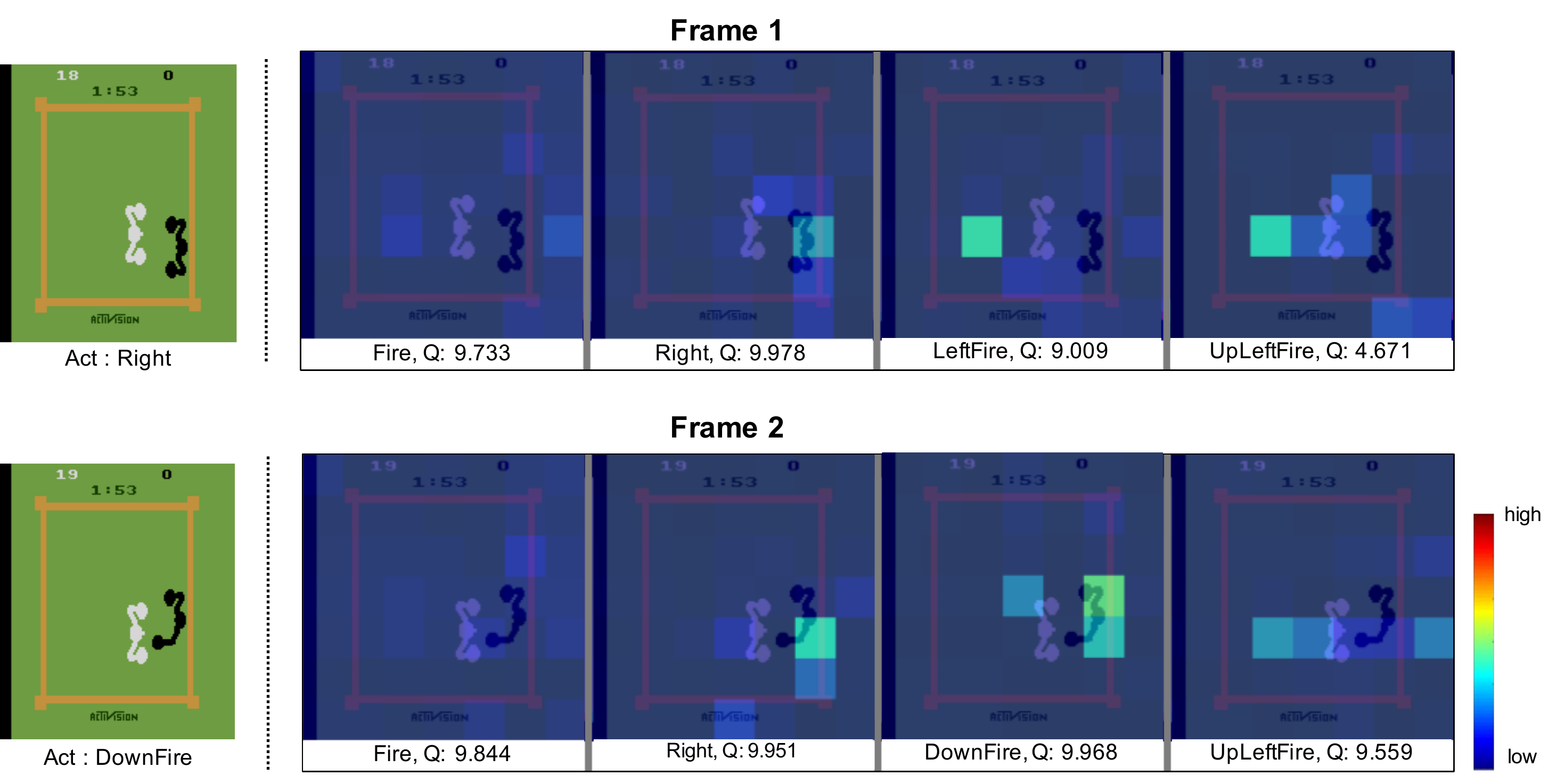}
\subcaption{\textbf{Boxing.}: 
The number of actions of Boxing’s agents is 18, but 4 actions are picked up and shown here.
Frame 1 is a scene where neither player's attacks reach each other, and Frame 2 is a scene where White can score points by attacking.
}
\label{fig:visual-boxing}
\end{minipage} \\
\end{tabular}
\caption{
\textbf{Visualization example of decoder-attention.}
The caption below the attention indicates the name of the action and the Q-value of the action. The caption at the bottom of the raw image indicates the action selected by the agent in the current frame. 
}
\label{fig:visual-decoder}
\end{figure}

{\bf Kung Fu Master.}

Kung Fu Master is a game in which the agent controls the player in the center of the screen and destroys enemies coming from the left and right sides of the screen.
The agent (=player) has 14 actions in total: 13 combinations of “Fire”, `“Up", “Down", “Left",“Right", and an additional “Noop".

From Frame 1, the agent is gazing at the enemy on the left for the action including ``Fire".
``Fire" is an attack action to destroy the enemy.
Therefore, we consider that the agent correctly recognizes the enemy as the target of the attack.

From Frame 2, the agent is gazing at the player's right side when the action includes ``Right" and at the player's left side when the action includes ``Left".
The ``Right" and ``Left" actions move the player in the specified direction.
Therefore, we can consider that the agent is gazing at the direction to which the player is moved by the action, and that the agent is correctly recognizing the action related to the movement.

{\bf Boxing.}

Boxing is a game in which the agent controls the white player to defeat the black player.
The agent (= white player) has 18 actions in total: 17 combinations of “Fire”, `“Up", “Down", “Left",“Right", and an additional “Noop".

From Frame 1, the agent is gazing at the white player's right side when the action includes ``Right" and at the white player's left side when the action includes ``Left".
The ``Right" and ``Left" actions move the white player to the specified direction.
Therefore, we can consider that the agent is gazing at the destination of the white player's move by the action and the agent is correctly aware of the action regarding the move.

From Frame 2, the agent is strongly gazing at the black player in ``DownFire'', which is an attack on the black player.
Also, ``Right" and ``Left" are the same as in Frame 1.
The agent's attack to the black player is directly related to the agent's score.
Therefore, we can consider that the agent recognizes the black player as the target of the attack and correctly recognizes the frames in which the black player can be attacked.

\end{document}